%% file: main.tex
\documentclass[letterpaper, 10 pt, conference]{ieeeconf}

\IEEEoverridecommandlockouts
\usepackage[T1]{fontenc}
\usepackage{cite}
\usepackage{url}
\usepackage{amsmath,amssymb,amsfonts}
\usepackage{graphicx}
\usepackage{textcomp}
\usepackage{xcolor}

\usepackage{booktabs}
\usepackage{multirow}
\usepackage{cuted}
\usepackage{needspace}
\usepackage[font=small]{caption}
\usepackage{tikz}
\makeatletter
\let\NAT@parse\undefined
\makeatother
\usepackage{hyperref}
\hypersetup{hidelinks}

\title{\LARGE\bfseries
NestDex: Nested Policy Learning with Copilot Assisted Teleoperation for Dexterous Manipulation
}

\author{
James Zhao$^{1,\dagger}$, Jinhe Tang$^{1,\dagger}$, Mingyuan Ba$^{1}$, and Weiming Zhi$^{1,2,3,*}$\\
\small $^{1}$School of Computer Science and $^{2}$Australian Centre for Robotics, The University of Sydney, Australia\\
\small $^{3}$College of Connected Computing, Vanderbilt University, TN, USA\\
\small $^{\dagger}$Equal contribution. $^{*}$Corresponding author: \texttt{Weiming.Zhi@sydney.edu.au}.\\
\small Project website: \href{https://aus.bot/research/nestdex}{\texttt{https://aus.bot/research/nestdex}}
}

\begin{document}

\raggedbottom

\maketitle
\thispagestyle{empty}
\pagestyle{empty}

\begin{abstract} Dexterous manipulation promises substantially richer robot interaction with the physical world, but learning these behaviours remains constrained by the difficulty of collecting consistent, complete-task demonstrations. Unlike parallel-jaw manipulation, dexterous tasks require the operator to coordinate arm motion with precise, contact-rich finger behaviour throughout the task. We introduce NestDex, a nested policy-learning framework that reduces this burden by using learned hand skills to assist demonstration collection. The operator controls the arm and regulates the active hand skill through a single-DoF clutch, rather than directly specifying the full finger trajectory. The inner hand policy adapts its motion from the latest proprioceptive history, while a vision-language selector activates the appropriate skill for each task stage. The resulting demonstrations train a separate outer visuomotor policy that controls both the arm and hand without the inner policies at deployment. A hand-action variational autoencoder provides compact hand-action targets while retaining arm commands in joint space. Across real-world dexterous manipulation experiments, NestDex improves demonstration reliability and efficiency, and the resulting empirical evaluations support effective autonomous policy learning.
\end{abstract}



\section{Introduction}

Imitation learning has produced capable visuomotor policies for robot manipulation, but their performance depends critically on consistent, complete-task demonstrations~\cite{ravichandar2020recent, Diff_templates,chi2023diffusion}. For dexterous manipulation, collecting such demonstrations is itself a major challenge. A parallel-jaw gripper typically reduces hand control to a single opening coordinate, whereas a multi-finger hand requires coordinated motion across many joints to establish, maintain, and adapt contact. Meanwhile, the operator must control the arm to progress through the task. Demonstration therefore requires simultaneously specifying \emph{where the arm should move} and \emph{how the hand should interact}, making data collection increasingly difficult as manipulation becomes more dexterous.

Existing dexterous teleoperation systems \cite{qin2023anyteleop} often still provide increasingly expressive control over the hand. The operator must still coordinate this control with arm motion throughout a multi-stage task. We instead consider a different division of labour: the operator guides task-level motion, while reusable hand skills generate the fine-grained contact behaviour required for demonstration. We introduce \textbf{NestDex}, which places learned hand policies inside the demonstration-collection loop. The operator controls the arm while a single-degree-of-freedom (DoF) clutch regulates a proprioceptive inner hand policy. The generated hand motion adapts from the latest proprioceptive history as contact evolves. The clutch is reversible, allowing the operator to advance or retract the generated hand motion, while a vision-language selector activates different inner policies as the task transitions between stages. Each demonstration therefore combines operator-directed arm motion with policy-generated dexterous hand behaviour, rather than requiring the operator to specify finger motions directly.

\begin{figure}[t]
    \centering
    \includegraphics[width=\linewidth]{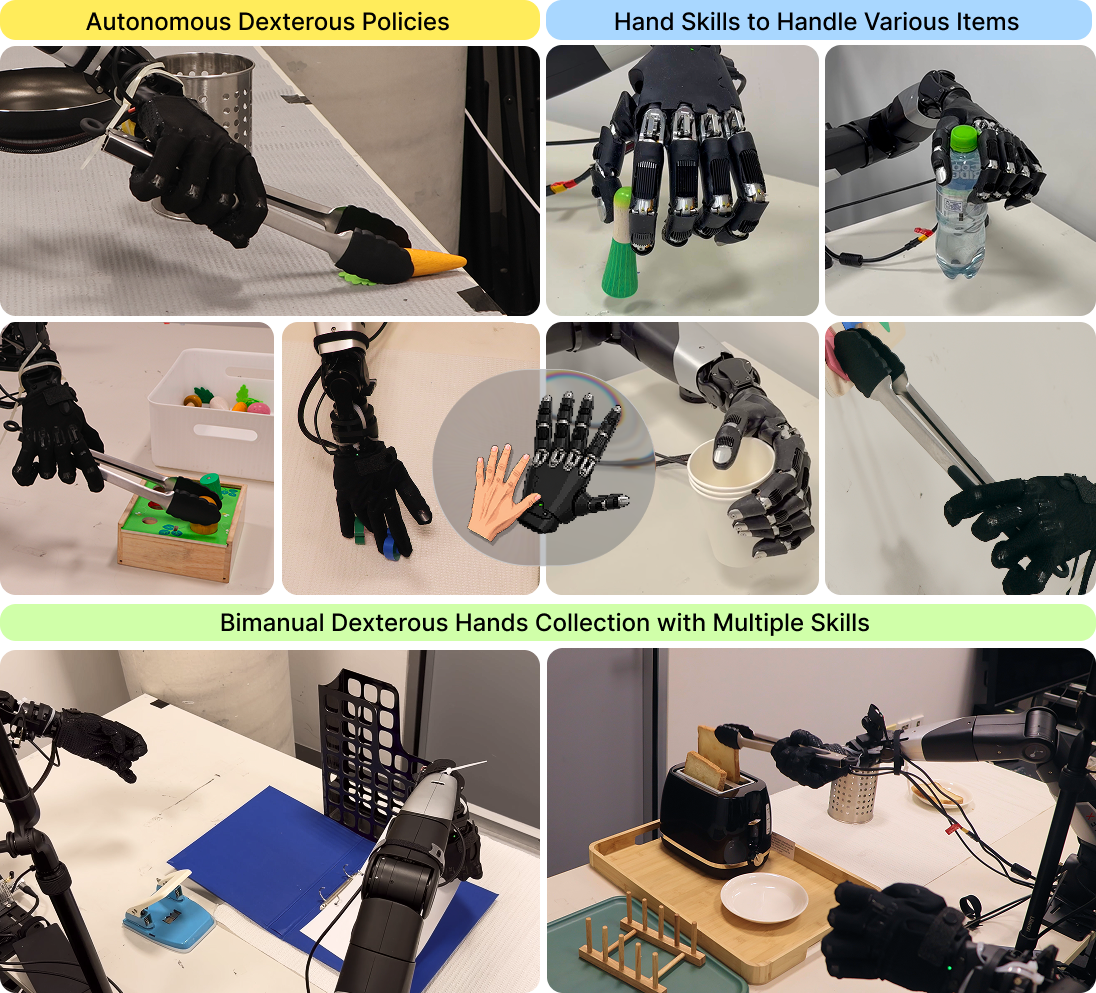}
    \caption{Overview of the capabilities evaluated with NestDex. Top left: autonomous outer-policy rollouts for Tongs Transfer and Dual-Object Transfer. Top right: one reusable inner policy produces coordinated hand configurations across varied items. Bottom: copilot assisted collection of bimanual Binder Filing and Toast Preparation demonstrations.}
    \label{fig:1}
\end{figure}

Importantly, these inner policies assist \emph{data collection}; they are not components of the final autonomous controller. The resulting demonstrations are then used to train a separate outer visuomotor policy that controls both the arm and hand at deployment. To make this high-dimensional action space easier to learn, a hand-action variational autoencoder (H-VAE) provides compact hand targets while retaining arm commands directly in joint space. We evaluate the resulting collection-to-autonomy pipeline from demonstration collection and autonomous policy learning through online execution, contact behaviour, and reuse of hand skills across tasks. Figure~\ref{fig:1} provides an overview of NestDex, along with example tasks.

\begin{figure*}[t]
    \centering
    \includegraphics[width=0.64\linewidth]{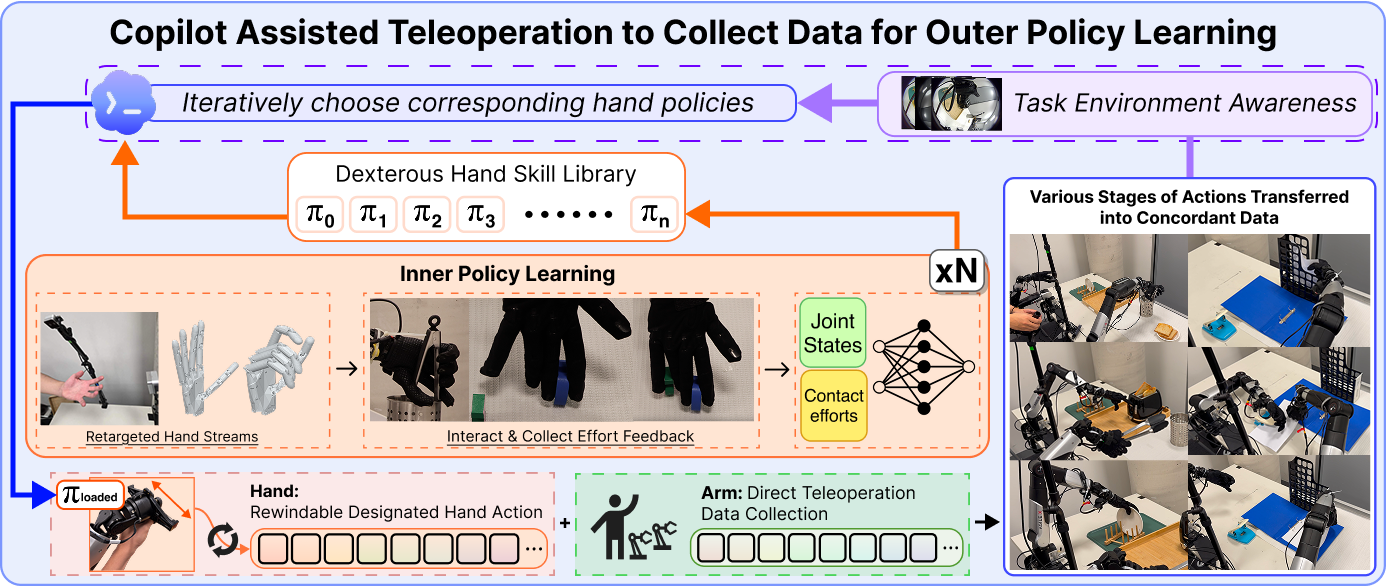}%
    \hfill
    \includegraphics[width=0.35\linewidth]{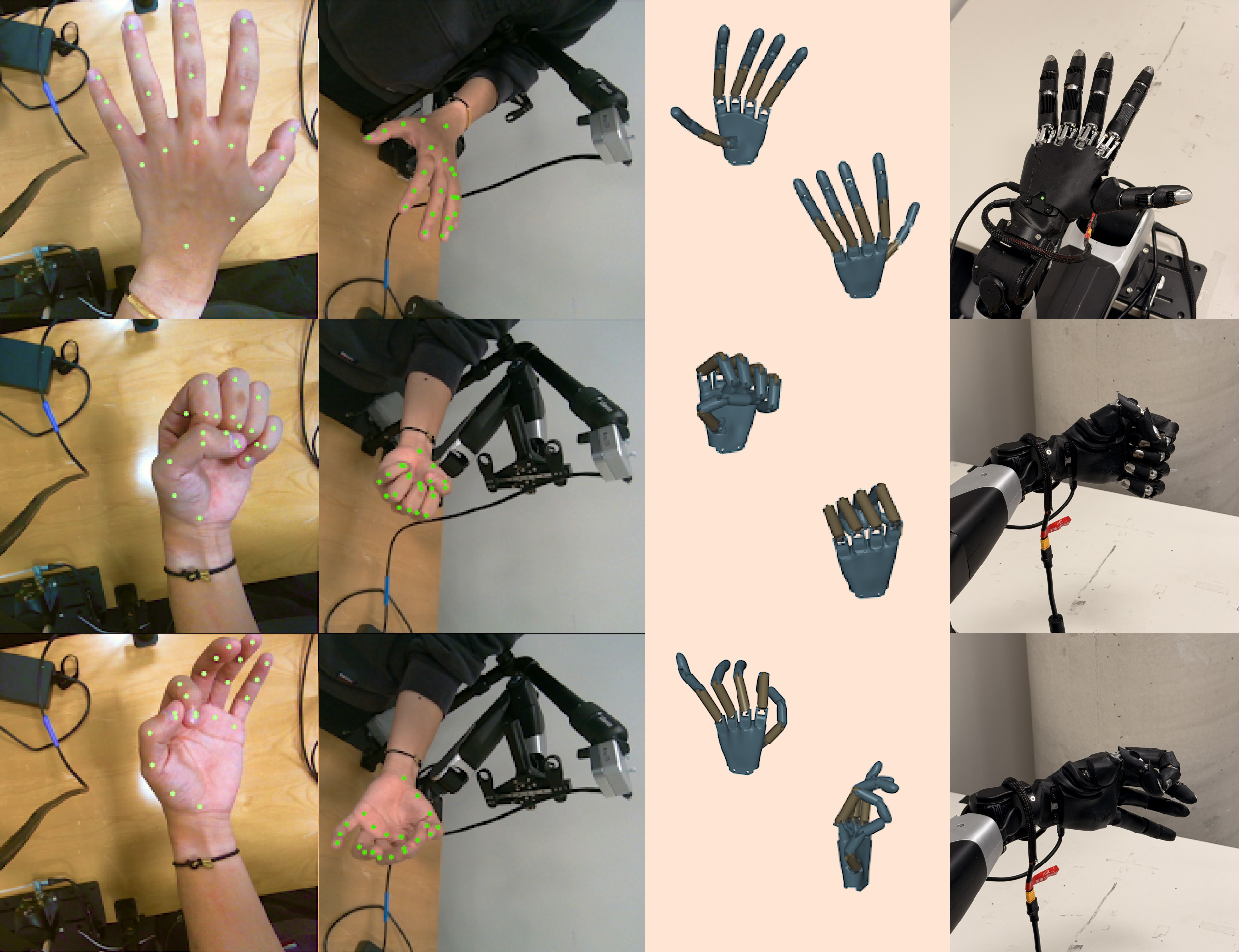}
    \caption{NestDex system overview and inner-policy data collection. Left: retargeted hand-skill demonstrations train inner policies that assist complete-task teleoperation; the collected demonstrations then train an outer policy. Right: multi-view human-hand observations are retargeted to coordinated robot-hand configurations for inner-policy training.}
    \label{fig:2}
\end{figure*}

Concretely, our technical contributions are:
\begin{itemize}
    \item \textbf{Copilot-assisted dexterous demonstration collection.} A reversible clutch allows an operator to regulate state-conditioned hand skills while controlling the arm, with automatic skill selection enabling different hand behaviours to be reused across task stages.
    \item \textbf{Nested collection and autonomous deployment.} Learned inner hand policies, dynamically allocated by a vision-language model, assist complete-task demonstration collection, while a separate outer visuomotor policy learns from the resulting demonstrations and executes independently at deployment.
    \item \textbf{Real-world collection-to-autonomy evaluation.} Across six dexterous manipulation tasks, we evaluate demonstration reliability and efficiency, downstream autonomous policy learning, and the effects of compact hand-action representations and online contact-aware execution.
\end{itemize}

\section{Related Work}

\noindent\textbf{Dexterous teleoperation and demonstration collection.}
Significant progress has been made in robotics to collect useful demonstration data to train policies \cite{li2026tripilotffcoordinatedwholebodyteleoperation, Post-Training}. Most dexterous demonstration systems map human hand motion to a robot hand. They capture finger motion with motion-capture gloves~\cite{rajeswaran2018dexterous,wang2024dexcap,zhang2025doglove} or estimate hand pose from camera observations~\cite{handa2020dexpilot,sivakumar2022telekinesis,qin2023anyteleop}. Specialised gloves can provide fine-grained hand demonstrations~\cite{zhang2025doglove}, but require additional hardware. In these systems, the operator provides the desired arm and hand motion throughout each demonstration. Shared autonomy instead allows a learned policy to control the hand while the operator directs the arm~\cite{cui2025sharedautonomy}. NestDex uses this division of control to collect complete-task demonstrations.

\vspace{0.5em}\noindent\textbf{Hierarchical and skill-based policy learning.}
Long-horizon robot tasks are often handled by breaking them into reusable skills. A high-level policy selects the next skill and may specify how it should be performed~\cite{sutton1999options,julian2020skills,lynch2020latentplans,pertsch2021spirl, GeometricFab}. Language-model agents can also select from an available skill set using task context~\cite{ichter2023saycan, barron2025cross}. These selectors and skills usually remain part of the deployed system. NestDex instead uses language-based selection during demonstration collection, then trains a separate policy to control the complete task without this component or the inner policies at deployment.

\vspace{0.5em}\noindent\textbf{Latent action representations for dexterous policy learning.}
The high-dimensional joint space of dexterous hands makes both exploration and supervised action prediction difficult. Compact action spaces have been used to constrain policy search in reinforcement learning~\cite{zhou2021plas,allshire2021laser,yang2024vqace}. Latent or discretised action representations have also been used to model complex, multimodal behaviour in imitation learning~\cite{shafiullah2022bet,lee2024vqbet,liconti2024latent}. NestDex applies a variational autoencoder~\cite{kingma2014vae} only to the recorded hand commands, using its lower-dimensional outputs as behaviour-cloning targets while retaining arm commands in joint space.

\section{Preliminaries}
\label{sec:preliminaries}

At time step \(t\), a visuomotor action chunk policy \(\pi_\theta\), parameterised by \(\theta\), receives a visual observation \(\mathbf{v}_t\) and a robot state \(\mathbf{s}_t\). Rather than predicting a single action, the policy predicts an action chunk \(\widehat{\mathbf{A}}_t\) with horizon \(H\):
\begin{equation}
    \label{eq:visuomotor_action_chunk}
    \widehat{\mathbf{A}}_t
    =\pi_{\theta}(\mathbf{v}_t,\mathbf{s}_t)
    =\left[\widehat{\mathbf{a}}_{t+1\mid t},\ldots,
    \widehat{\mathbf{a}}_{t+H\mid t}\right].
\end{equation}
Here, \(H\) is the action chunk horizon and \(\widehat{\mathbf{a}}_{k\mid t}\) denotes the action for time step \(k\) predicted using the observation available at time step \(t\). The same formulation applies to a proprioceptive action chunk policy. Given a history length \(h\), its input is the proprioceptive observation \(\mathbf{o}_t^{\mathrm{prop}}=[\mathbf{s}_{t-h+1},\ldots,\mathbf{s}_t]\), and its predicted chunk is \(\widehat{\mathbf{A}}_t=\pi_\theta(\mathbf{o}_t^{\mathrm{prop}})\). Predicting actions jointly captures short-term temporal dependencies and reduces the effective decision horizon~\cite{zhao2023act}.

Executing an entire action chunk before querying the policy again would delay the incorporation of new observations and could introduce discontinuities between consecutive chunks. Following prior work~\cite{zhao2023act}, the policy is instead queried at every time step, producing overlapping action chunks that contain multiple predictions for the same execution time. Let \(\mathcal{P}_t\) contain the predictions for time step \(t\), ordered from the oldest to the newest. The index \(i\in\{0,\ldots,|\mathcal{P}_t|-1\}\) identifies a prediction in this collection, where \(|\mathcal{P}_t|\) is the number of available predictions and \(\mathcal{P}_t[i]\) is the prediction at index \(i\). Temporal ensembling computes the executed action \(\mathbf{a}_t^{\mathrm{exec}}\) as
\begin{equation}
    \label{eq:temporal_ensembling}
    \mathbf{a}_t^{\mathrm{exec}}
    =\frac{\sum\nolimits_{i=0}^{|\mathcal{P}_t|-1}
    w_i\mathcal{P}_t[i]}
    {\sum\nolimits_{i=0}^{|\mathcal{P}_t|-1}w_i},
    \qquad w_i=\exp(-mi),
\end{equation}
Here, \(\mathcal{P}_t[0]\) is the oldest prediction and \(m>0\) controls the exponential weighting. A smaller \(m\) gives relatively more weight to predictions based on recent observations. By averaging predictions for the same execution step, temporal ensembling smooths chunk transitions while the policy continues to incorporate new observations.

\section{NestDex: Nested Dexterous Policies}
\label{sec:nestdex}
NestDex separates assisted demonstration collection from autonomous task execution (Figure~\ref{fig:2}). Retargeted hand-skill demonstrations train proprioceptive inner policies, which assist complete-task teleoperation. A variational autoencoder (VAE) for hand poses, which we call H-VAE, then encodes the collected hand commands for training an outer visuomotor policy that controls the complete task.

\textbf{Implementation.} Our experiments use a leader-follower platform. Each leader has an arm and a clutch, while each follower has an arm, a dexterous hand, and a wrist camera. Leader-arm joint positions map directly to follower-arm commands. The leader clutch regulates inner-policy execution on the follower hand, and each arm-hand pair executes its inner policy independently.

\subsection{Inner Policy Learning}

\subsubsection{Hand Skill Demonstration Collection}
Camera-based hand retargeting lets the operator collect multiple demonstrations of each dexterous hand skill through natural, coordinated finger movements.

First, we reconstruct the three-dimensional human-hand pose from multiple calibrated camera views. To reduce depth ambiguity and finger occlusion in single-view estimation, we triangulate synchronised hand keypoints across the camera views.

Second, we adopt the vector-based retargeting formulation of AnyTeleop~\cite{qin2023anyteleop} and replace its squared residual with a Huber penalty~\cite{huber1964robust} for robustness. At time step \(t\), the target joint configuration \(\mathbf{q}_t^{*}\) of the follower dexterous hand solves
\begin{equation}
\begin{aligned}
    \mathbf{q}_t^{*}
    = \underset{\mathbf{q}_{\min}\leq\mathbf{q}\leq\mathbf{q}_{\max}}
    {\arg\min}\quad
    &\sum_{i=1}^{M}
    \rho_{\delta}\!\left(
    \left\|\alpha\mathbf{v}_{i,t}^{h}
    -\mathbf{v}_i^{r}(\mathbf{q})\right\|_2\right) \\
    &+\beta\left\|\mathbf{q}-\mathbf{q}_{t-1}^{*}\right\|_2^2 .
\end{aligned}
\end{equation}
Here, \(\mathbf{q}\) denotes a candidate joint configuration of the follower dexterous hand within the joint limits \(\mathbf{q}_{\min}\) and \(\mathbf{q}_{\max}\), and \(\mathbf{q}_t^{*}\) is the optimal configuration at time step \(t\). For each of the \(M\) vector correspondences, \(\mathbf{v}_{i,t}^{h}\) is the human-hand vector obtained from the detected landmarks, while \(\mathbf{v}_i^{r}(\mathbf{q})\) is the corresponding robot-hand vector computed through forward kinematics. The Euclidean residual is evaluated using the Huber penalty \(\rho_{\delta}\) with threshold \(\delta\), which reduces sensitivity to large matching errors. The scale factor \(\alpha\) accounts for differences in hand size, and \(\beta\) penalises deviations from the previous solution \(\mathbf{q}_{t-1}^{*}\) to encourage temporal smoothness.

During retargeted execution, we record the follower hand's joint positions \(\mathbf{q}_{n,t}\) and joint effort \(\mathbf{e}_{n,t}\) at each time step \(t\). Each complete skill demonstration of length \(T_n\) forms a trajectory \(\tau_n=\{(\mathbf{q}_{n,t},\mathbf{e}_{n,t})\}_{t=0}^{T_n-1}\). For each skill, \(N\) trajectories constitute the inner-policy demonstration dataset \(\mathcal{D}_{\mathrm{in}}\).


\subsubsection{Inner Policy Training}
Given \(\mathcal{D}_{\mathrm{in}}\), we train one inner policy for each hand skill. We define the hand state at time step \(t\) as \(\mathbf{x}_{n,t}=[\mathbf{q}_{n,t},\mathbf{e}_{n,t}]\). Using a sliding window, each training sample pairs a proprioceptive observation containing the \(h\) most recent hand states with an action chunk containing the following \(H_{\mathrm{in}}\) joint positions from the same trajectory:
\begin{equation}
\begin{aligned}
    \mathbf{o}_{n,t}
    &= \left[\mathbf{x}_{n,t-h+1},\ldots,\mathbf{x}_{n,t}\right], \\
    \mathbf{A}_{n,t}
    &= \left[\mathbf{q}_{n,t+1},\ldots,
    \mathbf{q}_{n,t+H_{\mathrm{in}}}\right].
\end{aligned}
\end{equation}
Here, \(h\) is the observation history length and \(H_{\mathrm{in}}\) is the action chunk horizon. We extract every valid window, indexed by \(t\in\{h-1,\ldots,T_n-H_{\mathrm{in}}-1\}\), from all \(N\) trajectories to form the inner-policy training set \(\mathcal{S}_{\mathrm{in}}\).

Following the proprioceptive action chunk formulation in Section~\ref{sec:preliminaries}, we train the inner policy on \(\mathcal{S}_{\mathrm{in}}\) to map \(\mathbf{o}_{n,t}\) to \(\mathbf{A}_{n,t}\). The predicted chunks capture temporal coordination across the fingers and provide follower-hand joint-position commands during execution.

\begin{figure}[!t]
    \centering
    \includegraphics[width=\columnwidth]{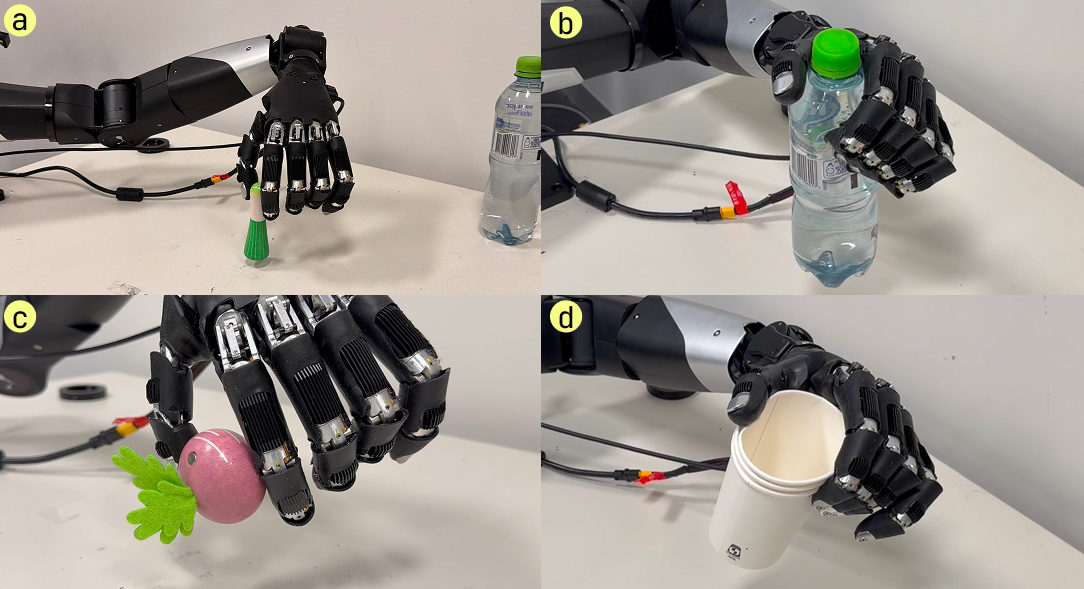}
    \caption{The inner policy produces different contact configurations across its four training objects: (a) green scallion toy, (b) water bottle, (c) radish toy, and (d) paper cup. The policy receives hand joint positions and efforts without object images or identities.}
    \label{fig:inner_policy_objects}
\end{figure}

For the grasp inner policy, \(\mathcal{D}_{\mathrm{in}}\) contains demonstrations collected with the four objects shown in Figure~\ref{fig:inner_policy_objects}: a green scallion toy, water bottle, rounded object, and paper cup. The demonstrations place the hand in different contact states while retaining the same grasp objective. One policy is trained across these trajectories, and its deployment input is the measured history of joint positions and efforts.

\subsection{Copilot Assisted Teleoperation}

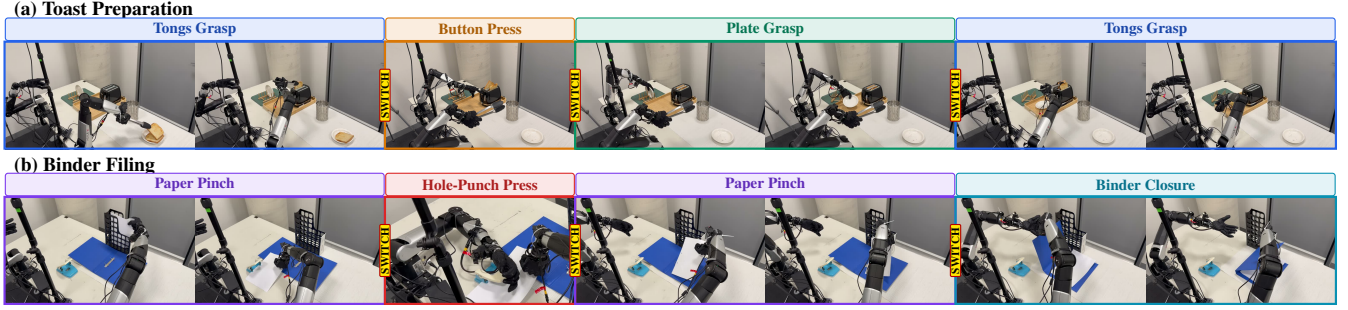
\begin{figure*}[!t]
    \centering
    \resizebox{\textwidth}{!}{%
        \input{figures/policy_switch_teleop}%
    }
    \caption{Copilot assisted teleoperation sequences for Toast Preparation and Binder Filing. Coloured borders and labels identify the active inner policy, while switch markers indicate automatic reselection after the preceding policy returns to zero execution index. Each sequence returns to a previously used policy when the corresponding hand skill is needed again.}
    \label{fig:policy_switch}
\end{figure*}

\textbf{Clutch guided inner-policy execution.} The operator teleoperates the follower arm and regulates follower-hand motion through the active inner policy. The leader-arm joint positions are mapped directly to follower-arm joint-position commands, while the leader clutch determines the active inner policy's \textbf{skill progress} \(p_t\in[0,1]\), a normalised scalar that specifies its target execution index. During each forward-execution control step, the inner policy is queried and its overlapping predictions are combined using the temporal ensembling rule in Eq.~\eqref{eq:temporal_ensembling}. For the demonstration dataset \(\mathcal{D}_{\mathrm{in}}\), the maximum trajectory length \(T_{\max}\) is defined as the length of its longest trajectory:
\begin{equation}
    T_{\max}=\max_{\tau_n\in\mathcal{D}_{\mathrm{in}}}T_n.
\end{equation}
Because all trajectories in \(\mathcal{D}_{\mathrm{in}}\) are successful skill executions, \(T_{\max}\) provides a conservative horizon that avoids truncating the skill before completion.
The clutch values assigned to the initial and terminal states of the dexterous-hand skill are \(c_{\mathrm{start}}\) and \(c_{\mathrm{end}}\), respectively. At time step \(t\), the current clutch input \(c_t\) is converted to skill progress \(p_t\), with the two endpoints corresponding to \(p_t=0\) and \(p_t=1\), respectively:
\begin{equation}
    p_t=\operatorname{clip}\!\left(
    \frac{c_t-c_{\mathrm{start}}}
    {c_{\mathrm{end}}-c_{\mathrm{start}}},0,1\right).
\end{equation}
The corresponding target execution index \(s_t^{\mathrm{in}}\) is
\begin{equation}
    s_t^{\mathrm{in}}=\left\lfloor p_t\cdot (T_{\max}-1)\right\rfloor,
    \qquad s_t^{\mathrm{in}}\in\{0,\ldots,T_{\max}-1\}.
\end{equation}

During teleoperation, \(s_t^{\mathrm{in}}\) is the target execution index specified by the clutch, whereas \(r_t^{\mathrm{in}}\) is the current execution index within the generated hand trajectory. The current execution index is initialised at zero and tracks the target index with a maximum change of one step per control cycle:
\begin{equation}
    r_{t+1}^{\mathrm{in}}
    =r_t^{\mathrm{in}}
    +\operatorname{clip}\!\left(
    s_t^{\mathrm{in}}-r_t^{\mathrm{in}},-1,1\right),
    \qquad r_0^{\mathrm{in}}=0.
\end{equation}

A positive index change queries the policy with the latest proprioceptive history, executes the next ensembled command, and appends it to the generated hand-trajectory buffer. Subsequent commands therefore reflect state changes produced by contact. A negative change executes the preceding buffered command and removes the abandoned command, while a zero change holds the current posture.

On reversal, we clear the accumulated ensemble predictions but retain the trajectory buffer, so predictions from the abandoned rollout do not affect forward execution after it resumes. The one-step index update prevents skipped commands, and the terminal index bounds execution at \(T_{\max}-1\). The operator can therefore reverse the hand to an earlier state and resume closed-loop prediction from the resulting proprioceptive history.

\vspace{0.5em}\noindent\textbf{Vision-language policy selection.} NestDex uses a pretrained vision-language agent to select the collection-time policy. The selector receives the wrist-camera image and a numbered list of inner-policy skill descriptions, then returns the policy index. Selection occurs at startup and whenever the operator fully reverses the active policy to \(r_t^{\mathrm{in}}=0\); the policy therefore cannot change during skill execution. The hand then moves to the selected policy's starting posture and initialises a new trajectory buffer. This routes each task stage to an inner policy while the operator controls arm motion and skill progress. Figure~\ref{fig:policy_switch} illustrates the resulting sequences in two tasks, which will be revisited in the empirical evaluations.

\subsection{Outer Policy Learning}

\begin{figure}[!t]
    \centering
    \includegraphics[width=\linewidth]{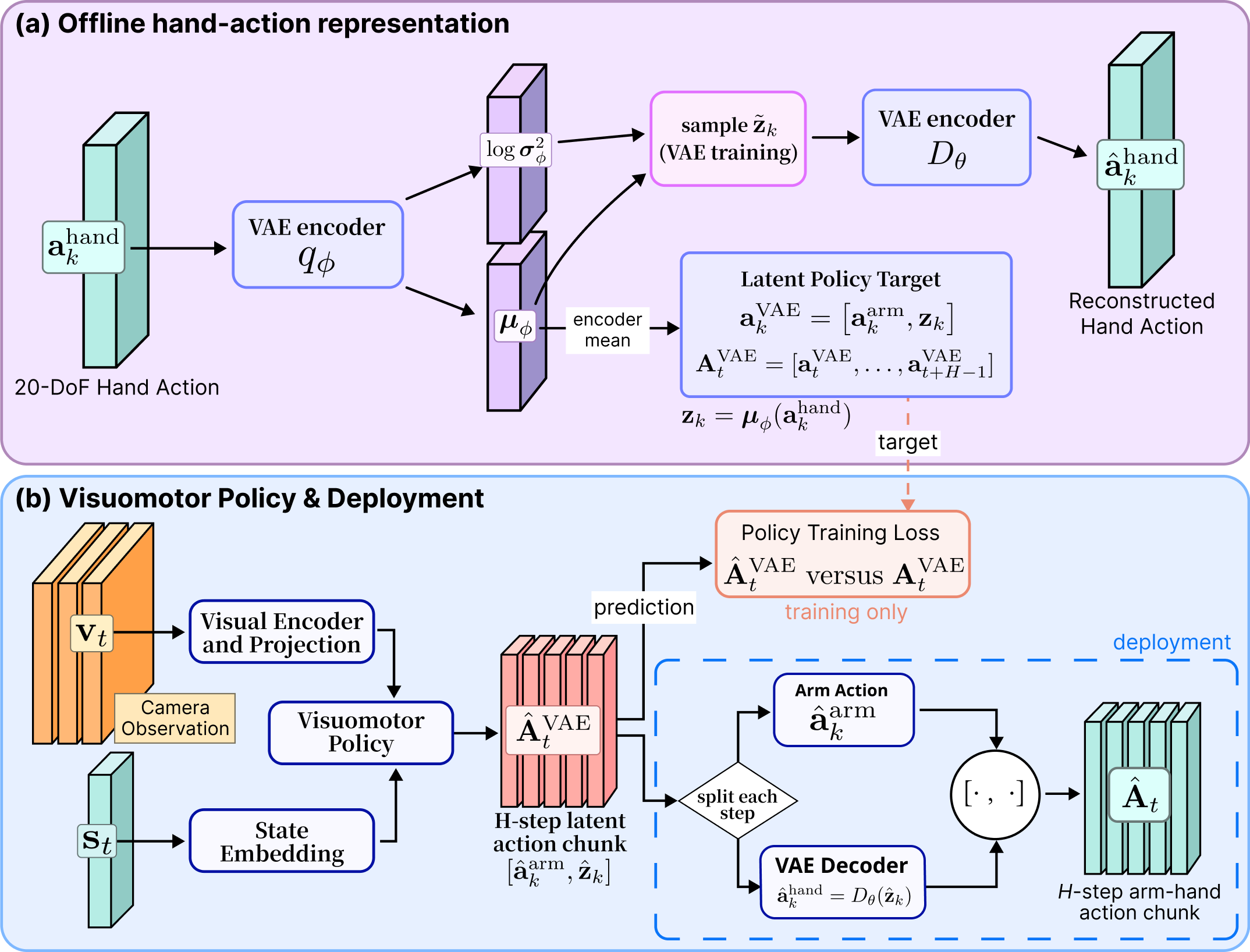}
    \caption{Hand action compression and outer-policy learning. (a) H-VAE encodes each 20-DoF hand joint-position command as a compact hand latent action and reconstructs the complete hand command through its decoder. (b) The outer policy predicts action chunks containing arm commands and hand latent actions; during deployment, the decoder converts each predicted hand latent action back to a follower-hand joint-position command.}
    \label{fig:outer_policy_learning}
\end{figure}

\begin{figure*}[!t]
    \centering
    \begin{minipage}[t]{0.49\textwidth}
        \centering
        \textbf{(a) Tongs Transfer}\par\vspace{2pt}
        \includegraphics[width=\linewidth]{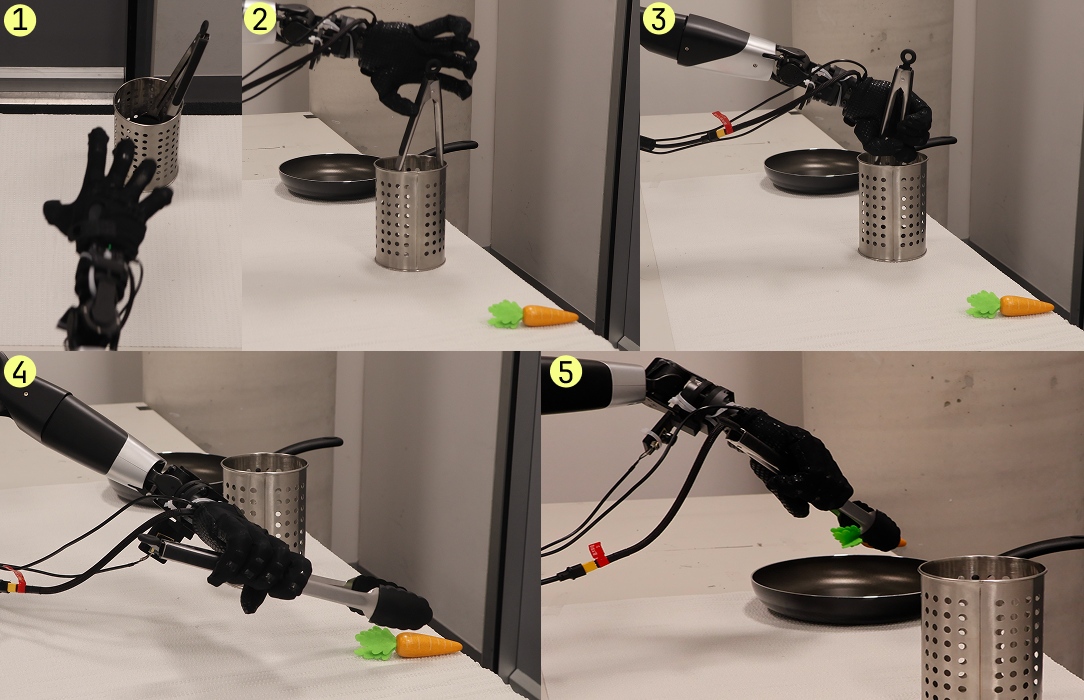}
    \end{minipage}
    \hfill
    \begin{minipage}[t]{0.49\textwidth}
        \centering
        \textbf{(b) Dual-Object Transfer}\par\vspace{2pt}
        \includegraphics[width=\linewidth]{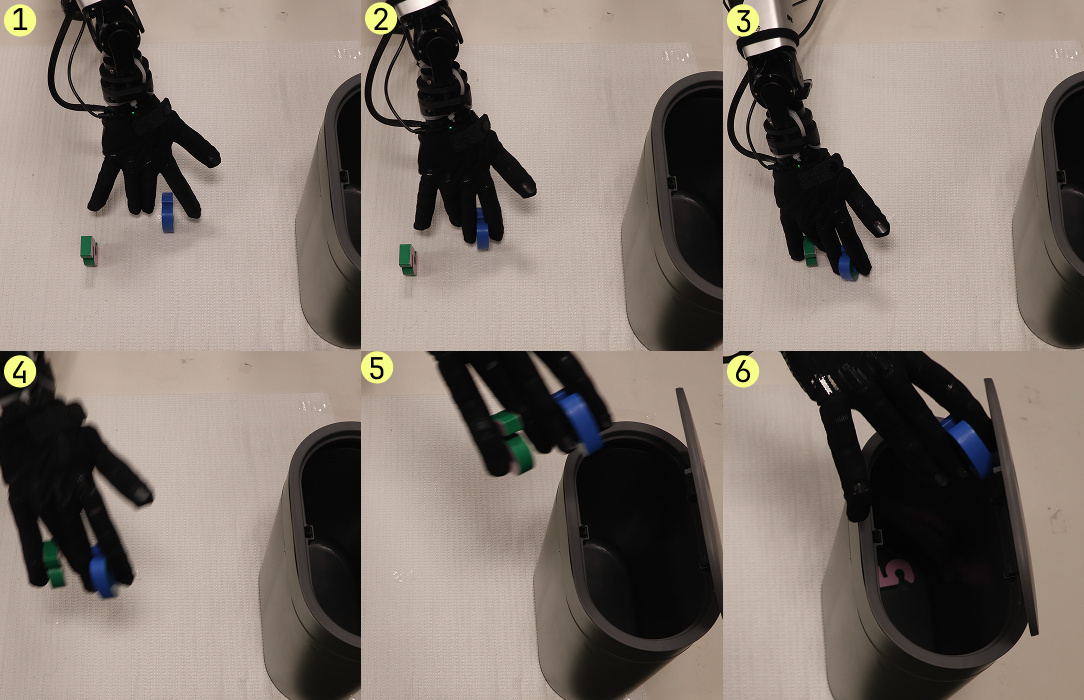}
    \end{minipage}

    \caption{Autonomous outer-policy behaviours learned from copilot-collected demonstrations. The policies deploy independently of the inner policies used during collection. (a) The policy extracts the tongs, grasps the carrot with the tool, and transfers it to the pan. (b) The policy uses separate finger groups to grasp two blocks simultaneously and transfers both to the bin. Numbers indicate temporal order.}
    \label{fig:autonomous_behaviours}
\end{figure*}

\subsubsection{Hand Variational Autoencoder Training}
Each complete task demonstration contains synchronised visual observations, measured arm and hand joint positions and efforts, and arm and hand joint-position commands. The outer policy operates on the arm-hand pairs required by each task. Each pair provides a wrist-camera image \(\mathbf{v}_t\), a state \(\mathbf{s}_t\) containing arm and hand joint positions and efforts, and the corresponding commands \(\mathbf{a}_t^{\mathrm{arm}}\) and \(\mathbf{a}_t^{\mathrm{hand}}\).

To reduce the output space of coordinated dexterous-hand motion, we use the H-VAE~\cite{kingma2014vae} to encode the hand joint-position command as \(\mathbf{z}_k\), while leaving the arm commands unchanged. We follow the setup in the original variation Autoencoder paper ~\cite{kingma2014vae}, and train it with reconstruction and Kullback--Leibler (KL) regularisation using reparameterisation. After training, we use the posterior mean \(\overline{\mathbf{z}}_k=\boldsymbol{\mu}_{\phi}(\mathbf{a}_k^{\mathrm{hand}})\), where \(\phi\) denotes the encoder parameters, to avoid sampling noise in the outer-policy targets. Each outer-policy action label \(\mathbf{a}_k\) and action chunk \(\mathbf{A}_t\) are then defined as
\begin{equation}
\begin{aligned}
    \mathbf{a}_k
    &=\left[\mathbf{a}_k^{\mathrm{arm}},\overline{\mathbf{z}}_k\right], \\
    \mathbf{A}_t
    &=\left[\mathbf{a}_{t+1},\ldots,\mathbf{a}_{t+H}\right].
\end{aligned}
\end{equation}
Here, \(H\) is the outer-policy action chunk horizon.
The outer-policy training set \(\mathcal{D}_{\mathrm{out}}\) comprises samples \((\mathbf{v}_t,\mathbf{s}_t,\mathbf{A}_t)\) extracted from the complete task demonstrations.
\subsubsection{Outer Policy Training}
Following the visuomotor action chunk formulation in Eq.~\eqref{eq:visuomotor_action_chunk}, the outer policy maps \((\mathbf{v}_t,\mathbf{s}_t)\) to \(\mathbf{A}_t\). Using \(\mathcal{D}_{\mathrm{out}}\), we train this policy by behaviour cloning~\cite{pomerleau1988alvinn}. The behaviour cloning objective is
\begin{equation}
\mathcal{L}_{\mathrm{BC}}(\theta)
=\mathbb{E}_{\mathcal{D}_{\mathrm{out}}}
\left[\ell\!\left(\pi_{\theta}(\mathbf{v}_t,\mathbf{s}_t),
\mathbf{A}_t\right)\right],
\end{equation}
where \(\theta\) parameterises the policy and \(\ell\) is its supervised action prediction loss. During inference, the outer policy is queried at every control step and its overlapping action chunk predictions are combined using Eq.~\eqref{eq:temporal_ensembling}. The resulting arm joint-position command is sent to the follower arm, while the resulting hand latent action is decoded by the H-VAE into a follower-hand joint-position command.

Figure~\ref{fig:outer_policy_learning} summarises the H-VAE and its integration with outer-policy training and deployment.

\begin{table}[t]
\centering
\caption{Definitions and manipulation modes of the six evaluated dexterous manipulation tasks.}
\label{tab:task_definitions}
\footnotesize
\setlength{\tabcolsep}{3pt}
\renewcommand{\arraystretch}{1.12}
\begin{tabular}{p{0.26\columnwidth} p{0.16\columnwidth} p{0.46\columnwidth}}
\toprule
\textbf{Task} & \textbf{Mode} & \textbf{Description} \\
\midrule
Tongs Transfer
& Single-arm
& Extract the tongs, use them to grasp a wooden carrot, and place the carrot into the pan. \\
Bottle Disposal
& Single-arm
& Open the bin lid, grasp an empty plastic bottle, and place it inside. \\
Dual-Object Transfer
& Single-arm
& Grasp two wooden number blocks simultaneously and place them into the bin. \\
Ingredient and Pot Transfer
& Single-arm
& Grasp a wooden scallion and place it into the pot, then grasp the pot and move it to the target position. \\
Toast Preparation
& Dual-arm
& Use the tongs to place the toast in the toaster, press the heating button, position the plate, and return the heated toast to the plate. \\
Binder Filing
& Dual-arm
& Remove paper from the rack, punch it, insert it into the blue ring binder, and close the rings. \\
\bottomrule
\end{tabular}
\end{table}

\begin{table*}[t]
\centering
\begin{minipage}[t]{0.59\textwidth}
\centering
\caption{Same-platform comparison of demonstration collection across six tasks.}
\label{tab:olift_six_tasks}
\resizebox{\linewidth}{!}{%
\begin{tabular}{llcccccc}
\toprule
Method & Metric & \shortstack{Tongs\\Transfer} & \shortstack{Bottle\\Disposal} & \shortstack{Dual-Object\\Transfer} & \shortstack{Ingredient and\\Pot Transfer} & \shortstack{Toast\\Preparation} & \shortstack{Binder\\Filing} \\
\midrule
\multirow{2}{*}{Copilot}
& Time (s) & 44.33 & 41.37 & 36.19 & 43.26 & 327.46 & 221.80 \\
& Success Rate (\%) & 100\% & 100\% & 100\% & 100\% & 100\% & 100\% \\
\multirow{2}{*}{AnyTeleop}
& Time (s) & N/A & 88.88 & 121.63 & 55.29 & N/A & N/A \\
& Success Rate (\%) & 0\% & 50\% & 30\% & 75\% & 0\% & 0\% \\
\bottomrule
\end{tabular}%
}
\end{minipage}\hfill
\begin{minipage}[t]{0.395\textwidth}
\centering
\caption{Outer-policy results across four tasks for different demonstrations.}
\label{tab:olift_ours_act}
\resizebox{\linewidth}{!}{%
\begin{tabular}{lcccc}
\toprule
Method & \shortstack{Tongs\\Transfer} & \shortstack{Bottle\\Disposal} & \shortstack{Dual-Object\\Transfer} & \shortstack{Ingredient and\\Pot Transfer} \\
\midrule
Copilot, no H-VAE & 65\% & 60\% & 80\% & 85\% \\
Copilot, H-VAE & 100\% & 75\% & 90\% & 100\% \\
AnyTeleop, no H-VAE & N/A & 40\% & 20\% & 75\% \\
\bottomrule
\end{tabular}%
}
\end{minipage}
\end{table*}

\section{Empirical Evaluation}

We structure the evaluation around six questions: \textbf{Q1}, the cost and reliability of demonstration collection; \textbf{Q2}, whether copilot demonstrations support autonomous outer-policy learning; \textbf{Q3}, the benefit of the H-VAE hand-action representation; \textbf{Q4}, online execution and temporal ensembling; \textbf{Q5}, how behaviour varies across training objects; and \textbf{Q6}, policy selection and reuse across task stages.

\begin{figure}[t]
    \centering
    \includegraphics[width=\columnwidth]{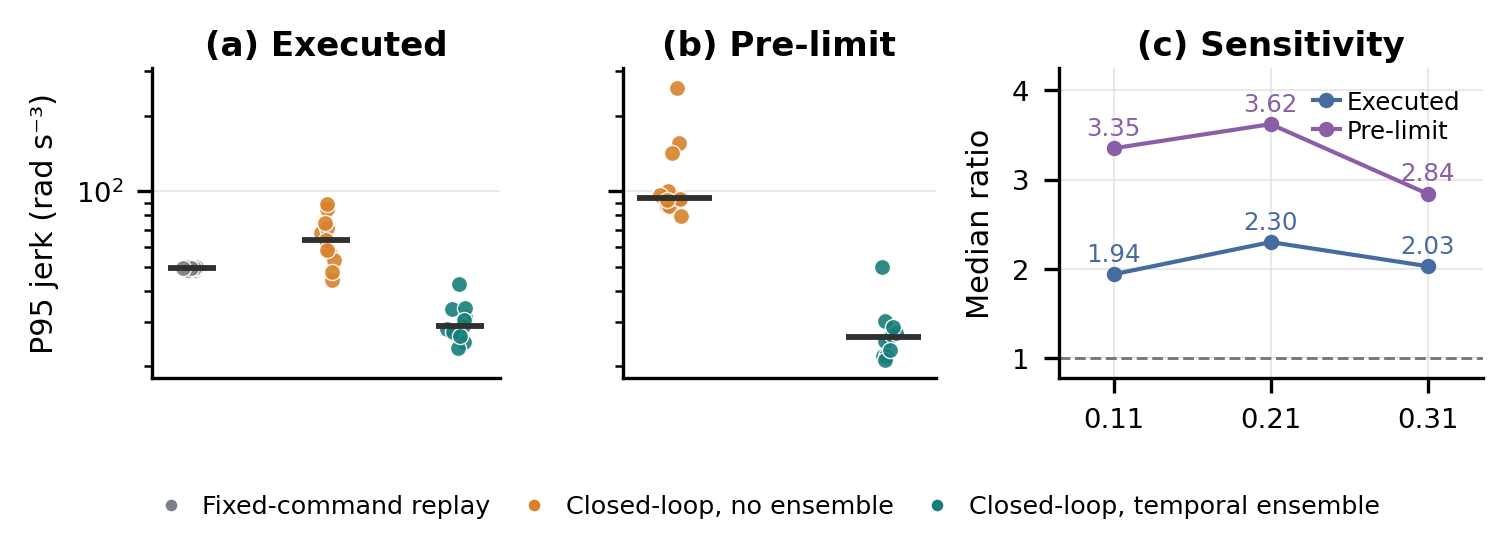}
    \caption{Temporal ensembling ablation during bottle-grasp closure. Panels (a) and (b) show trial-level executed-command and pre-limit P95 jerk; black bars mark medians. Panel (c) shows the median ratio of Closed-loop, no ensemble to Closed-loop, temporal ensemble across filter windows. Lower values indicate smoother commands.}
    \label{fig:temporal_ensembling_jerk_ablation}
\end{figure}

\begin{figure}[t]
    \centering
    \vspace{0.5em}
    \includegraphics[width=0.95\columnwidth]{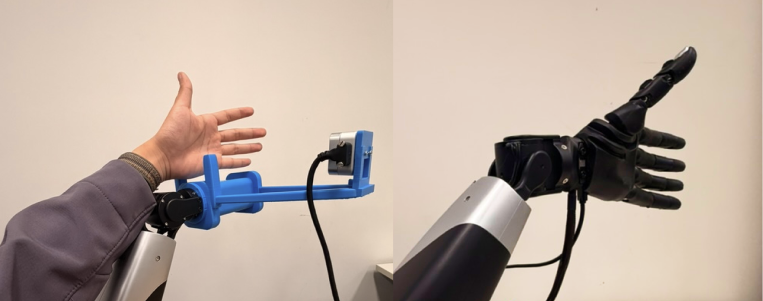}
    \caption{AnyTeleop~\cite{qin2023anyteleop} baseline used for demonstration collection. The operator moves the leader arm while presenting the hand to the mounted observation camera (left); AnyTeleop maps the observed hand pose to the follower dexterous hand (right).}
    \label{fig:anyteleop_setup}
\end{figure}

\vspace{0.5em}\noindent\textbf{Physical setup.}
All experiments ran on the leader-follower platform described in \hyperref[sec:nestdex]{Section~\ref*{sec:nestdex}}. Each arm-hand pair paired a leader side with a 7-DoF Piper Nero arm and a 1-DoF clutch against a follower side with a matching 7-DoF arm, a 20-DoF five-finger WujiHand I, and a wrist camera. At every time step the follower hand state comprised a 20-dimensional joint-position vector and a 20-dimensional joint-effort vector, while the follower arm state comprised a 7-dimensional joint-position vector and a 7-dimensional joint-effort vector. Inner-policy execution ran at 100~Hz. Figure~\ref{fig:anyteleop_setup} shows the AnyTeleop baseline used in Q1. Running all baselines on this identical setup eliminates hardware and kinematic confounders to strictly isolate the impact of copilot-assisted hand control.

\begin{figure*}[t]
    \centering
    \includegraphics[width=0.95\textwidth]{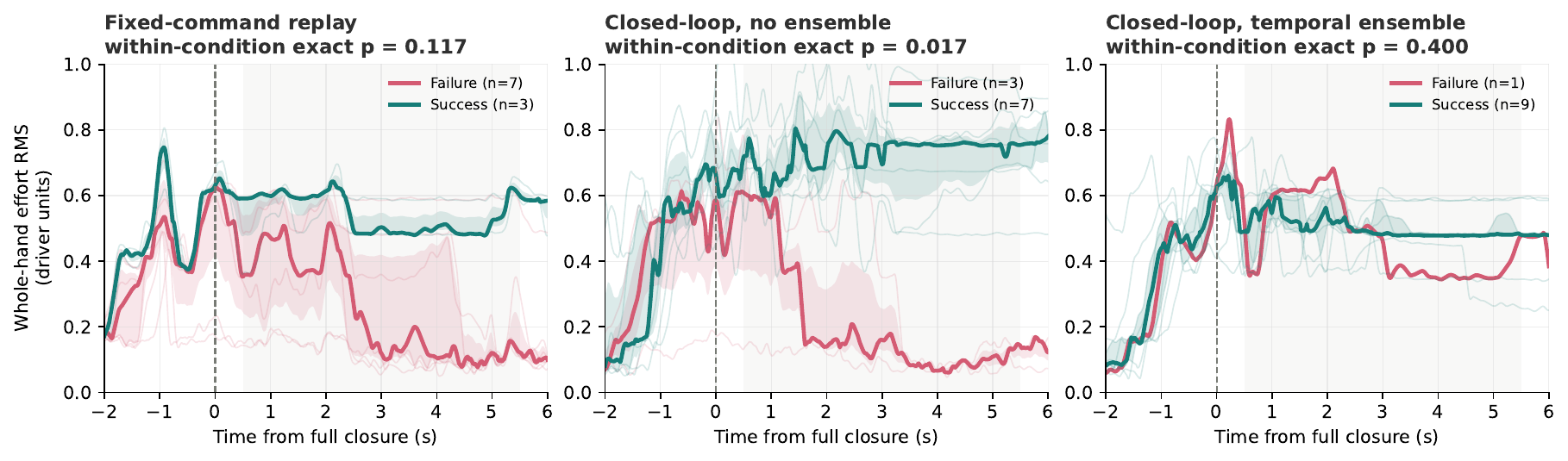}
    \caption{Closure-aligned whole-hand effort across the 30 bottle-grasp trials. Successes and failures remain separate within each execution condition. Thin curves show trials, thick curves show the median, and shaded regions show the interquartile range when more than one trial is available. The grey interval marks 0.5 to 5.5~s after full closure.}
    \label{fig:effort_across_execution_curves}
\end{figure*}

\begin{figure}[t]
    \centering
    \includegraphics[width=\columnwidth]{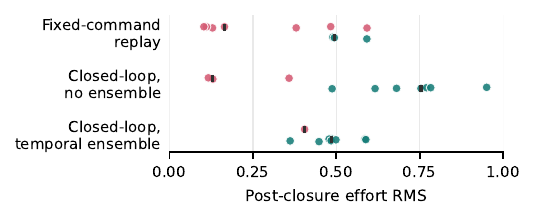}
    \caption{Trial-level post-closure effort from 0.5 to 5.5~s after closure. Points show trial medians; pink denotes failure, teal success, and black marks group medians.}
    \label{fig:effort_across_execution_trials}
\end{figure}

\vspace{0.5em}\noindent\textbf{Training setup.} Each inner policy used a proprioception-only Transformer (four encoder layers and one decoder layer) to map 30-step hand-state histories to 30-step action chunks, trained on 10 trajectories with AdamW for 20,000 steps (batch size 256, learning rate $10^{-5}$). The H-VAE~\cite{kingma2014vae} used $[128,64]$ hidden layers to map standardised 20-dimensional hand commands to 10-dimensional latent actions, and was trained for 100 epochs (batch size 512, learning rate $10^{-3}$). The outer policy employed a visuomotor Transformer (four encoder layers and one decoder layer) with visual embeddings from DINOv3 pretrained on LVD-1689M~\cite{simeoni2025dinov3} to predict 100-step action chunks from $256\times256$ RGB images, and was trained with AdamW for 50,000 steps (batch size 8, learning rate $10^{-5}$).

\vspace{0.5em}\noindent\textbf{Metrics.}
A teleoperation attempt or policy rollout is considered successful only when it completes every steps in the corresponding task description in Table~\ref{tab:task_definitions}. The success rate is the percentage of all attempts or rollouts that are successful.
For demonstration collection, the reported time per successful demonstration is the total duration of all successful and failed task attempts divided by the number of successful demonstrations. For NestDex, this metric also includes the one-time cost of collecting the 10 trajectories used to train each required inner policy, amortised across the successful complete-task demonstrations. The metric is undefined when no successful demonstration is obtained.

\subsection{Reliable and Efficient Demonstration Collection (Q1)}

We compared copilot-assisted teleoperation against AnyTeleop across six dexterous manipulation tasks, collecting twenty demonstrations per task with each system. Success rate and time per successful demonstration, defined under Metrics, are reported in Table~\ref{tab:olift_six_tasks}. N/A indicates that no successful demonstration was obtained. The same operator performed both conditions after familiarisation with each interface, with objects reset to the same nominal starting configurations between attempts.

\vspace{0.5em}\noindent\textbf{Results.}
As shown in Table~\ref{tab:olift_six_tasks}, NestDex achieves higher demonstration-collection success rates than AnyTeleop across the evaluated tasks and a lower time per successful demonstration on all tasks for which a direct comparison is possible. It also yields successful demonstrations on tasks for which AnyTeleop produces no successful demonstrations. We attribute these gains to the separation of arm teleoperation from dexterous-hand motion generation. The follower arm is teleoperated through direct leader-to-follower joint-position mapping, while inner-policy execution for the follower hand is regulated through the leader clutch. This decomposition removes high-dimensional finger coordination from direct arm-hand teleoperation and provides temporally coordinated hand commands, improving the reliability and efficiency of demonstration collection.

\subsection{From Copilot to Autonomy (Q2 and Q3)}

\noindent\textbf{Shared protocol.}
For Q1, we evaluate 20 task attempts per method. For Q2, on tasks where a method produced successful demonstrations, we continued collection as needed to obtain 20 successful complete-task trajectories for outer-policy training. The demonstration source is the sole experimental difference in the comparison between teleoperation systems. We also compare direct hand joint-position actions with latent actions from an H-VAE trained separately for each task. Each policy is evaluated over 20 rollouts, with task success reported in Table~\ref{tab:olift_ours_act}. The N/A entry for Tongs Transfer follows from AnyTeleop's 0\% demonstration success in Table~\ref{tab:olift_six_tasks}.

\vspace{0.5em}\noindent\textbf{Results (Q2).}
Table~\ref{tab:olift_ours_act} shows that outer policies trained on copilot-collected demonstrations achieve effective success rates across all evaluated tasks, confirming the utility of these demonstrations for downstream policy learning. With direct hand actions, policies trained on copilot-collected demonstrations outperform those trained on AnyTeleop demonstrations on every task for which both sources provide training data. For Tongs Transfer, AnyTeleop does not provide successful demonstrations for outer-policy training, whereas copilot-collected demonstrations support successful autonomous task execution.

\noindent\textbf{Results (Q3).}
As shown in Table~\ref{tab:olift_ours_act}, incorporating the H-VAE consistently improves success rates across all four tasks, indicating that hand latent actions facilitate outer-policy learning. Hand joint-position commands exhibit strong coordination structure, so directly predicting these high-dimensional commands increases output complexity and can encourage the policy to fit local variations in the demonstrations. The H-VAE instead maps the correlated commands to compact hand latent actions, allowing the outer policy to focus on task-relevant hand coordination patterns while the decoder reconstructs the complete follower-hand joint-position commands. This reduced hand action space makes prediction more tractable and improves outer-policy learning from limited demonstrations. Figure~\ref{fig:autonomous_behaviours} provides qualitative rollout examples of the resulting outer policies trained on copilot-collected demonstrations: one performs a multi-stage tool transfer, while the other coordinates separate finger groups to grasp two objects simultaneously.

\subsection{Online Execution and Contact Behaviour (Q4 and Q5)}
\label{sec:inner_policy_execution_ablation}

\noindent\textbf{Protocol (Q4).}
We test the same inner policy on a water-bottle grasp under three execution conditions. Fixed-command replay repeats the joint-command trajectory saved from one successful rollout. Closed-loop, no ensemble queries the policy from the latest proprioceptive history, but executes the first action of each new chunk. Closed-loop, temporal ensemble applies Eq.~\eqref{eq:temporal_ensembling} to overlapping predictions from the same policy. For a balanced comparison, we use the first 10 trials with an adjudicated outcome from each condition, giving 30 trials in total.

For trajectory analysis, we isolate the active closing phase from 5\% to 95\% of the command-space displacement towards the closed posture. Every trajectory is resampled at 100~Hz. We compute the 95th percentile of absolute joint jerk after a third-order Savitzky-Golay filter with a 0.21~s window, both before the per-step command limit and after the command is delivered to the hand. Figure~\ref{fig:temporal_ensembling_jerk_ablation} also reports the result with 0.11~s and 0.31~s windows. For the effort diagnostic, Figures~\ref{fig:effort_across_execution_curves} and~\ref{fig:effort_across_execution_trials} align each trial to full closure and report whole-hand effort RMS in driver units.

\vspace{0.5em}\noindent\textbf{Results (Q4).}
The results separate two benefits of online inner-policy execution. First, recomputing actions from the latest proprioceptive history improves robustness to contact variation: Fixed-command replay succeeds in 3/10 trials, while Closed-loop, no ensemble and Closed-loop, temporal ensemble succeed in 7/10 and 9/10 trials, respectively. The difference between Fixed-command replay and Closed-loop, temporal ensemble is significant ($p=0.0198$), while the present sample does not establish a success-rate difference between the two closed-loop conditions. This suggests that a hand trajectory that succeeds once can be brittle when replayed open loop, whereas closed-loop prediction can respond to the latest proprioceptive history as contact evolves.

Second, temporal ensembling substantially smooths this closed-loop behaviour. Closed-loop, no ensemble has $2.30\times$ greater executed-command P95 jerk (the 95th percentile of the magnitude of jerk) than Closed-loop, temporal ensemble ($p=1.8\times10^{-4}$), with the same trend before command limiting and across all tested filter windows, without increasing closing duration. The contact-effort traces provide a consistent diagnostic: successful Closed-loop, no ensemble trials sustain greater post-closure effort than successful Closed-loop, temporal ensemble trials, suggesting more abrupt contact interaction. Together, these results indicate complementary roles: closed-loop prediction provides contact-dependent adaptation, while temporal ensembling makes that adaptation substantially smoother.

\noindent\textbf{Results (Q5).}
The same grasp policy produces distinct hand configurations across the four objects represented in its training data (Figure~\ref{fig:inner_policy_objects}). Importantly, the policy receives neither images nor object identities; its only inputs are the evolving joint positions and efforts. Object geometry is therefore reflected indirectly through contact with the hand, allowing the generated motion to evolve differently as different constraints are encountered. This behaviour is consistent with a contact-conditioned hand skill rather than replay of a fixed finger trajectory. Because all four objects are represented during training, this experiment demonstrates adaptation across learned contact conditions rather than generalisation to unseen objects.

\begin{figure}[t]
    \centering
    \includegraphics[width=1\linewidth]{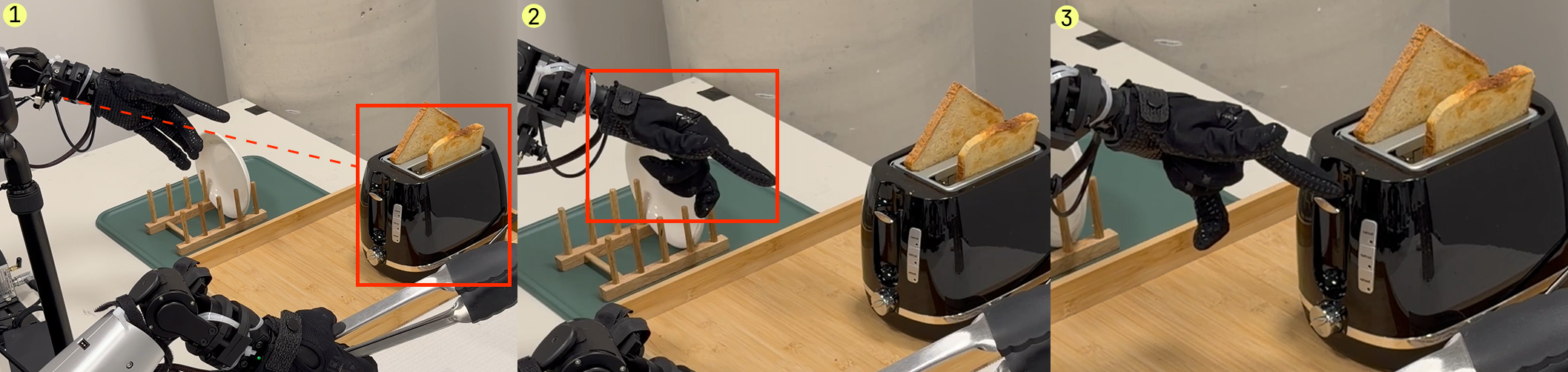}
    \caption{Automatic inner-policy switching during Toast Preparation. (1) Wrist-camera input selects Button Press. (2) The hand moves to the skill's initial posture. (3) Clutch input executes the button press.}
    \label{fig:toast_policy_switching}
\end{figure}

\subsection{Skill Selection and Reuse Across Task Stages (Q6)}
\noindent\textbf{Protocol (Q6).}
We qualitatively examine inner-policy selections during representative Toast Preparation and Binder Filing demonstrations. Figure~\ref{fig:toast_policy_switching} shows the Toast Preparation sequence.

\Needspace{5\baselineskip}
\vspace{0.5em}\noindent\textbf{Results (Q6).}
In both tasks, the selected skill follows the current stage, and previously used skills are reactivated when needed. Toast Preparation selects Tongs Grasp, Button Press, and Plate Grasp before returning to Tongs Grasp. Binder Filing returns to Paper Pinch after Hole-Punch Press, then selects Binder Closure. These sequences show that the selector reuses a compact set of hand skills across task stages.

\section{Conclusion}
Dexterous manipulation is often limited less by policy capacity than by the difficulty of collecting reliable demonstrations that coordinate arm motion with complex hand behaviour. NestDex addresses this bottleneck by placing reusable, state-conditioned hand skills inside the demonstration process, allowing the operator to focus on task-level motion while the hand policy manages fine-grained dexterous coordination. Across real-world dexterous manipulation experiments, NestDex improves demonstration reliability and efficiency, and its collected demonstrations train autonomous arm-hand policies that successfully execute the evaluated tasks. A compact hand-action representation further improves downstream learning, while the bottle-grasp study shows the importance of retaining online, contact-aware execution and smoothing its predictions through temporal ensembling. Overall, NestDex provides a practical route from assisted dexterous demonstration collection to fully autonomous manipulation, while preserving the adaptability needed for contact-rich hand behaviour.

\bibliographystyle{IEEEtran}
\bibliography{reference}

\end{document}

%% file: figures/policy_switch_teleop.tex
\definecolor{tongpolicy}{HTML}{2563EB}%
\definecolor{buttonpolicy}{HTML}{D97706}%
\definecolor{platepolicy}{HTML}{059669}%
\definecolor{paperpolicy}{HTML}{7C3AED}%
\definecolor{punchpolicy}{HTML}{DC2626}%
\definecolor{ringpolicy}{HTML}{0891B2}%
\begin{tikzpicture}[x=1cm,y=1cm]
    \tikzset{
        tasklabel/.style={
            anchor=west,
            font=\fontsize{7.2}{7.8}\selectfont\bfseries
        },
        policybar/.style={
            rounded corners=1pt,
            minimum height=0.29cm,
            inner sep=0pt,
            font=\fontsize{5.6}{6.1}\selectfont\bfseries
        },
        switchlabel/.style={
            fill=yellow!88!orange,
            draw=red!70!black,
            text=black,
            rounded corners=0.5pt,
            inner xsep=1.1pt,
            inner ysep=0.45pt,
            rotate=90,
            font=\fontsize{4.1}{4.4}\selectfont\bfseries
        }
    }

    \def\framewidth{2.45}
    \def\frameheight{1.378}

    \node[tasklabel] at (0,3.16) {(a) Toast Preparation};

    \node[policybar,draw=tongpolicy,fill=tongpolicy!12,text=tongpolicy!72!black,
          minimum width=4.88cm] at (2.45,2.92) {Tongs Grasp};
    \node[policybar,draw=buttonpolicy,fill=buttonpolicy!13,text=buttonpolicy!78!black,
          minimum width=2.43cm] at (6.125,2.92) {Button Press};
    \node[policybar,draw=platepolicy,fill=platepolicy!12,text=platepolicy!78!black,
          minimum width=4.88cm] at (9.80,2.92) {Plate Grasp};
    \node[policybar,draw=tongpolicy,fill=tongpolicy!12,text=tongpolicy!72!black,
          minimum width=4.88cm] at (14.70,2.92) {Tongs Grasp};

    \node[anchor=north west,inner sep=0pt] at (0,2.76)
        {\includegraphics[width=\framewidth cm,height=\frameheight cm]{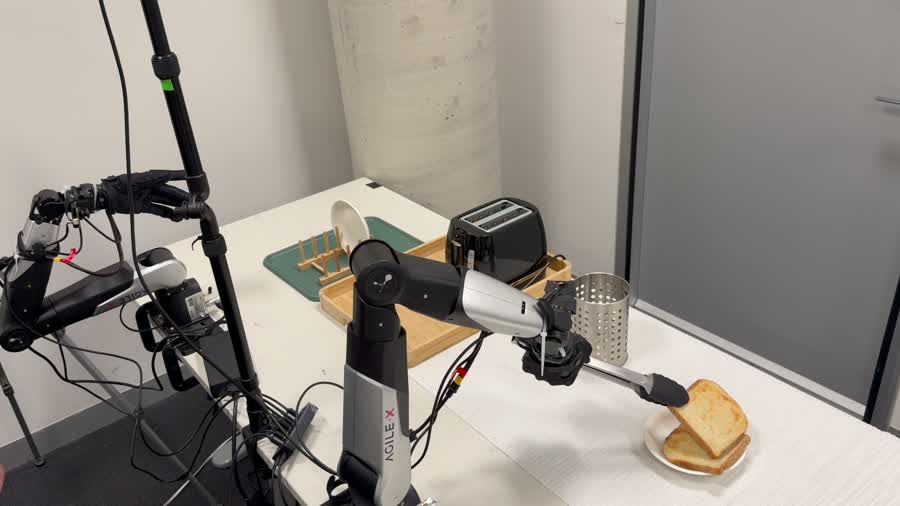}};
    \node[anchor=north west,inner sep=0pt] at (2.45,2.76)
        {\includegraphics[width=\framewidth cm,height=\frameheight cm]{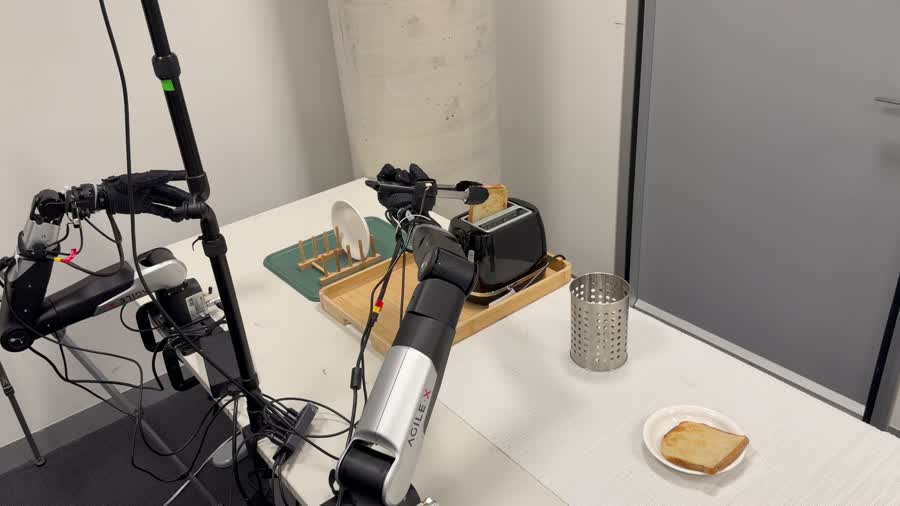}};
    \node[anchor=north west,inner sep=0pt] at (4.90,2.76)
        {\includegraphics[width=\framewidth cm,height=\frameheight cm]{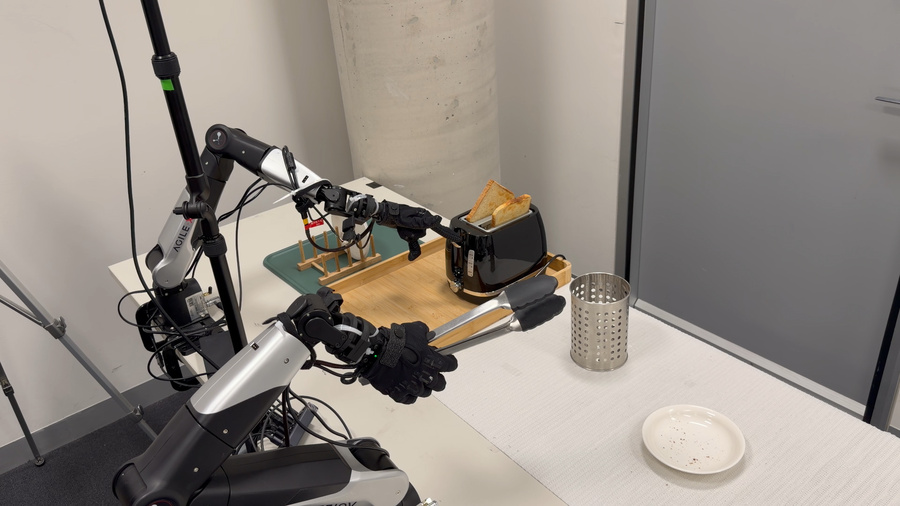}};
    \node[anchor=north west,inner sep=0pt] at (7.35,2.76)
        {\includegraphics[width=\framewidth cm,height=\frameheight cm]{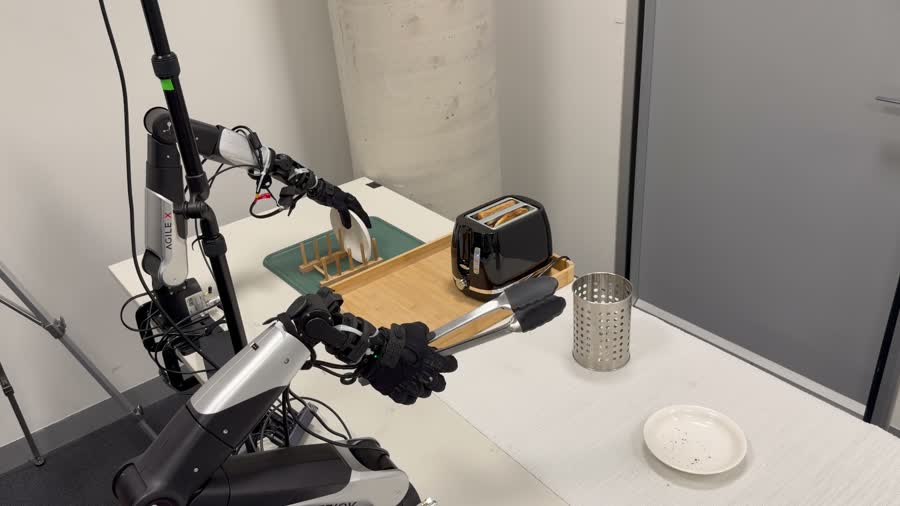}};
    \node[anchor=north west,inner sep=0pt] at (9.80,2.76)
        {\includegraphics[width=\framewidth cm,height=\frameheight cm]{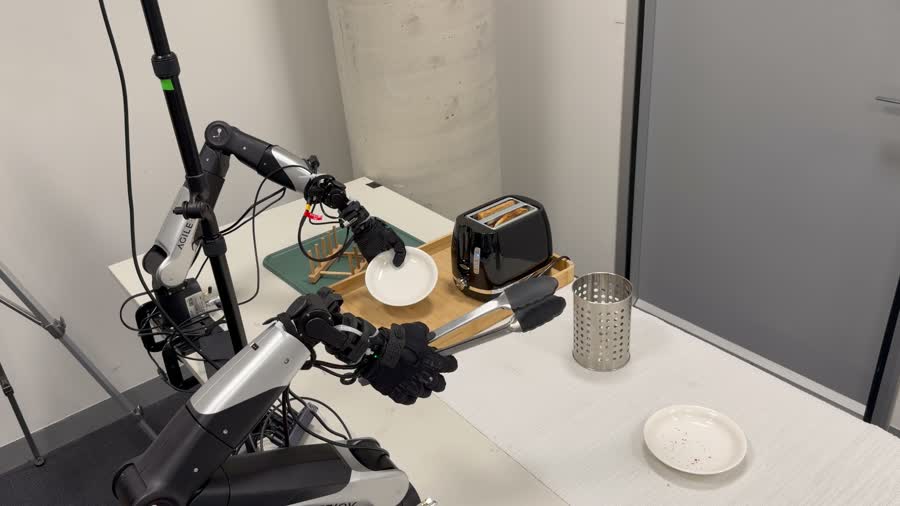}};
    \node[anchor=north west,inner sep=0pt] at (12.25,2.76)
        {\includegraphics[width=\framewidth cm,height=\frameheight cm]{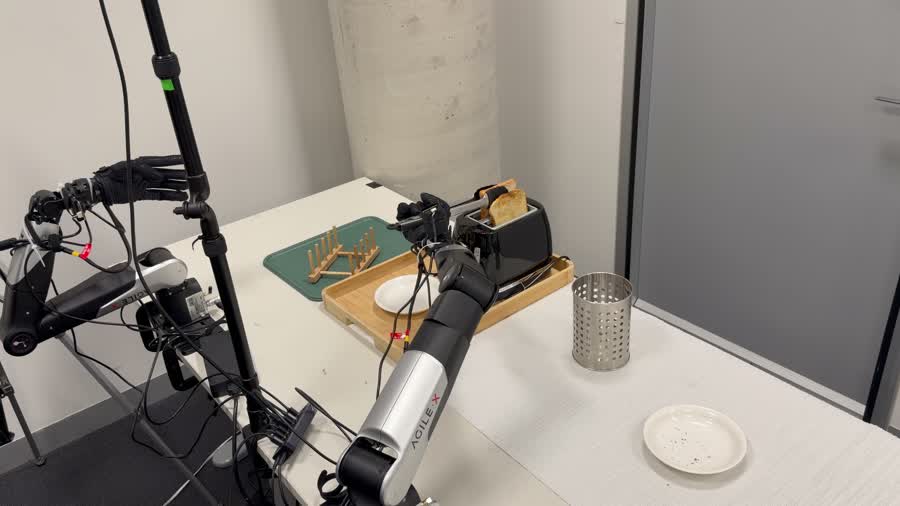}};
    \node[anchor=north west,inner sep=0pt] at (14.70,2.76)
        {\includegraphics[width=\framewidth cm,height=\frameheight cm]{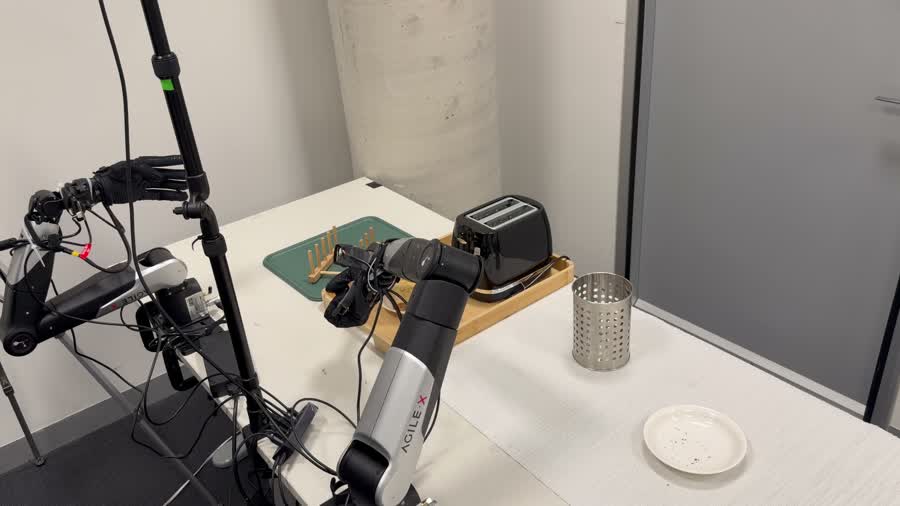}};

    \draw[tongpolicy,line width=0.8pt] (0,1.382) rectangle (4.90,2.76);
    \draw[buttonpolicy,line width=0.8pt] (4.90,1.382) rectangle (7.35,2.76);
    \draw[platepolicy,line width=0.8pt] (7.35,1.382) rectangle (12.25,2.76);
    \draw[tongpolicy,line width=0.8pt] (12.25,1.382) rectangle (17.15,2.76);

    \node[switchlabel] at (4.90,2.07) {SWITCH};
    \node[switchlabel] at (7.35,2.07) {SWITCH};
    \node[switchlabel] at (12.25,2.07) {SWITCH};

    \node[tasklabel] at (0,1.16) {(b) Binder Filing};

    \node[policybar,draw=paperpolicy,fill=paperpolicy!11,text=paperpolicy!78!black,
          minimum width=4.88cm] at (2.45,0.92) {Paper Pinch};
    \node[policybar,draw=punchpolicy,fill=punchpolicy!11,text=punchpolicy!78!black,
          minimum width=2.43cm] at (6.125,0.92) {Hole-Punch Press};
    \node[policybar,draw=paperpolicy,fill=paperpolicy!11,text=paperpolicy!78!black,
          minimum width=4.88cm] at (9.80,0.92) {Paper Pinch};
    \node[policybar,draw=ringpolicy,fill=ringpolicy!11,text=ringpolicy!78!black,
          minimum width=4.88cm] at (14.70,0.92) {Binder Closure};

    \node[anchor=north west,inner sep=0pt] at (0,0.76)
        {\includegraphics[width=\framewidth cm,height=\frameheight cm]{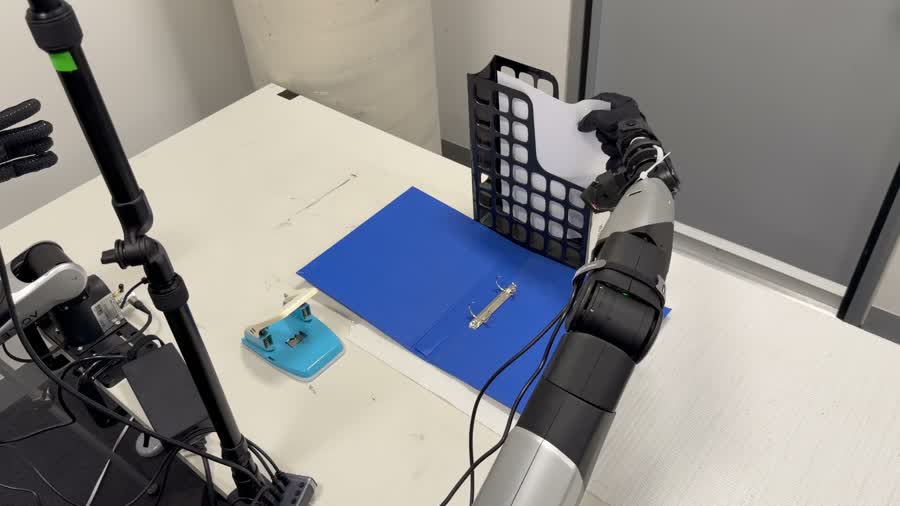}};
    \node[anchor=north west,inner sep=0pt] at (2.45,0.76)
        {\includegraphics[width=\framewidth cm,height=\frameheight cm]{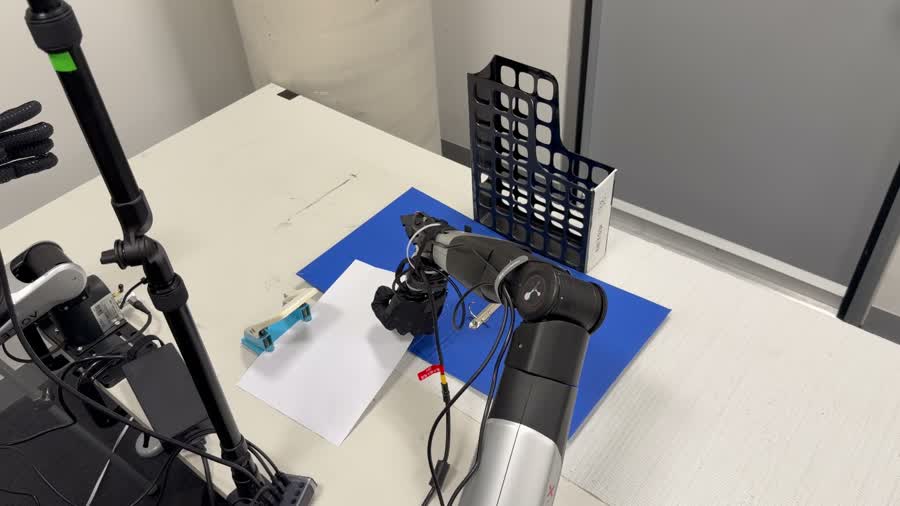}};
    \node[anchor=north west,inner sep=0pt] at (4.90,0.76)
        {\includegraphics[width=\framewidth cm,height=\frameheight cm]{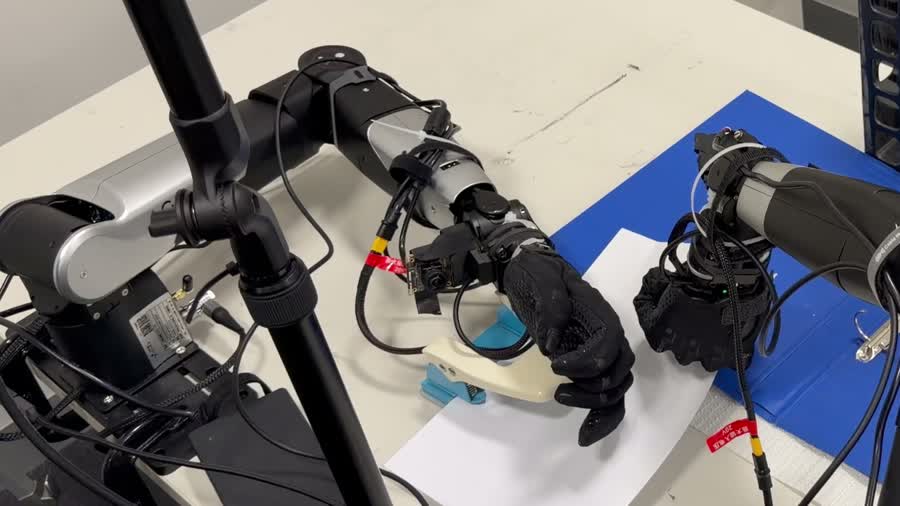}};
    \node[anchor=north west,inner sep=0pt] at (7.35,0.76)
        {\includegraphics[width=\framewidth cm,height=\frameheight cm]{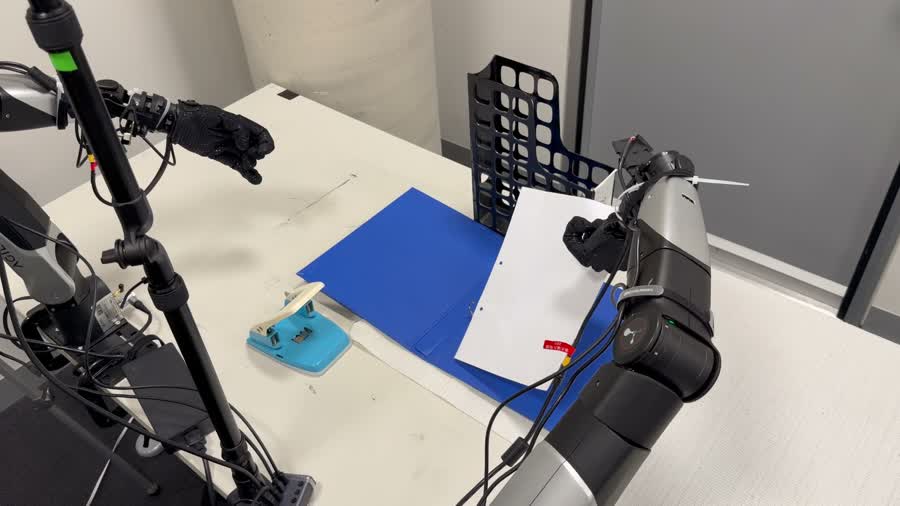}};
    \node[anchor=north west,inner sep=0pt] at (9.80,0.76)
        {\includegraphics[width=\framewidth cm,height=\frameheight cm]{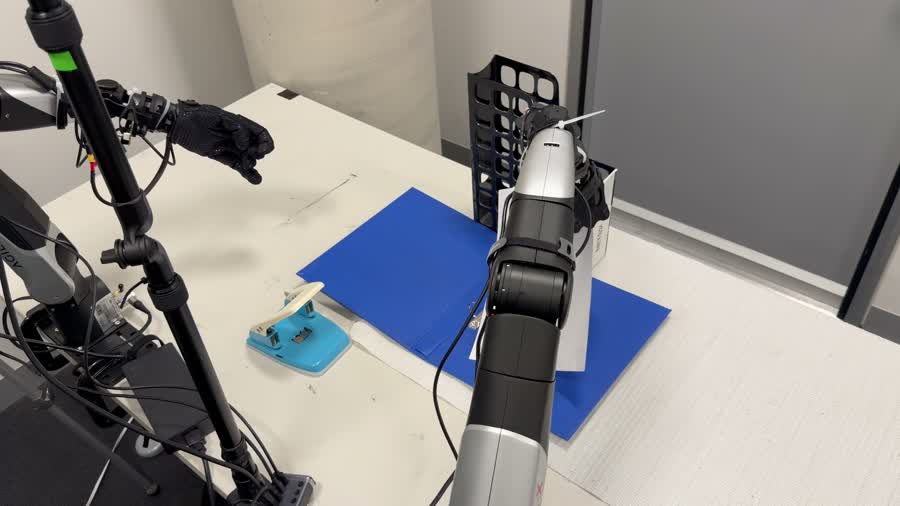}};
    \node[anchor=north west,inner sep=0pt] at (12.25,0.76)
        {\includegraphics[width=\framewidth cm,height=\frameheight cm]{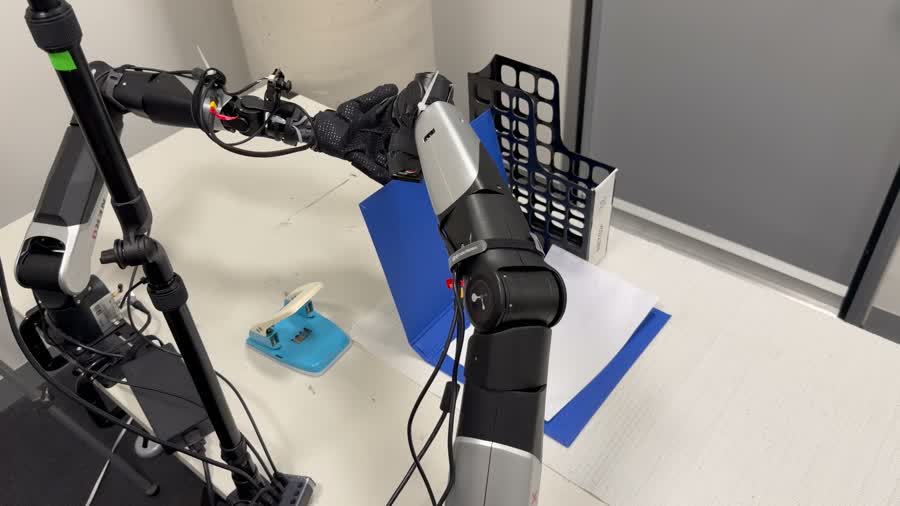}};
    \node[anchor=north west,inner sep=0pt] at (14.70,0.76)
        {\includegraphics[width=\framewidth cm,height=\frameheight cm]{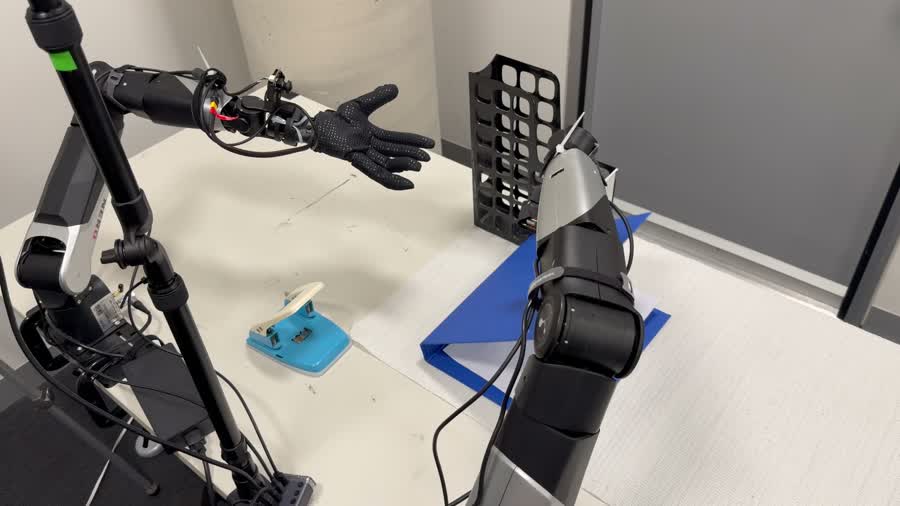}};

    \draw[paperpolicy,line width=0.8pt] (0,-0.618) rectangle (4.90,0.76);
    \draw[punchpolicy,line width=0.8pt] (4.90,-0.618) rectangle (7.35,0.76);
    \draw[paperpolicy,line width=0.8pt] (7.35,-0.618) rectangle (12.25,0.76);
    \draw[ringpolicy,line width=0.8pt] (12.25,-0.618) rectangle (17.15,0.76);

    \node[switchlabel] at (4.90,0.07) {SWITCH};
    \node[switchlabel] at (7.35,0.07) {SWITCH};
    \node[switchlabel] at (12.25,0.07) {SWITCH};

    \pgfresetboundingbox
    \path[use as bounding box] (-0.02,-0.64) rectangle (17.17,3.38);
\end{tikzpicture}